\documentclass[pdflatex,sn-basic,Numbered]{sn-jnl}

\usepackage{graphicx}
\usepackage{booktabs}
\usepackage{multirow}
\usepackage{amsmath,amssymb}
\usepackage{array}
\usepackage{xcolor}
\usepackage{textcomp}
\usepackage{url}
\usepackage{listings}
\usepackage{longtable}

\definecolor{promptdark}{HTML}{1F4E79}
\definecolor{promptlight}{HTML}{EDF4FA}
\lstdefinestyle{promptstyle}{
  basicstyle=\ttfamily\footnotesize,
  backgroundcolor=\color{promptlight},
  frame=single,
  rulecolor=\color{promptdark},
  framerule=0.55pt,
  framesep=5pt,
  xleftmargin=5.55pt,
  xrightmargin=5.55pt,
  breaklines=true,
  breakatwhitespace=false,
  breakindent=0pt,
  breakautoindent=false,
  columns=fullflexible,
  keepspaces=true,
  showstringspaces=false,
  upquote=true,
  aboveskip=0pt,
  belowskip=8pt
}

\newcommand{\promptheading}[1]{%
  \par\addvspace{6pt}\noindent
  \begingroup\setlength{\fboxsep}{4.5pt}%
  \colorbox{promptdark}{\parbox{\dimexpr\linewidth-2\fboxsep\relax}{%
    \color{white}\normalfont\bfseries\footnotesize\MakeUppercase{#1}}}%
  \endgroup\par\nobreak\vspace{-0.55pt}\nobreak
}

\begin{document}

\title[Bactrainus for multi-hop question answering]{Bactrainus: Optimizing Large Language Models for Multi-hop Complex Question Answering Tasks}

\author[1]{\fnm{Iman} \sur{Barati}}\email{iman\_barati@comp.iust.ac.ir}
\author[1]{\fnm{Arash} \sur{Ghafouri}}\email{aghafuri@comp.iust.ac.ir}
\author*[1]{\fnm{Behrouz} \sur{Minaei-Bidgoli}}\email{b\_minaei@iust.ac.ir}

\affil*[1]{\orgdiv{School of Computer Engineering}, \orgname{Iran University of Science and Technology}, \orgaddress{\city{Tehran}, \country{Iran}}}

\abstract{Multi-hop question answering requires a system to identify and integrate evidence distributed across documents, yet large language models remain vulnerable to irrelevant context. We investigate this evidence bottleneck in the English HotpotQA distractor setting and introduce Bactrainus, a modular selector--reader framework that separates paragraph selection, supporting-sentence identification, and answer generation. Optional question decomposition and teacher-generated rationale supervision make it possible to test where additional reasoning structure is useful. The evaluation combines foundation-model screening, controlled context and prompting ablations, parameter-efficient adaptation of Llama~3.1~8B Instruct and Llama~3.1~70B Instruct readers, and integrated selector--reader experiments. Supplying the full candidate context instead of gold supporting facts reduces answer token-overlap F1 by 17--21 points, showing that scale alone does not remove context sensitivity. The largest observed differences are associated with reader adaptation and sentence-level evidence control. The strongest reported configuration obtains 89.01 answer F1 and 79.70 joint F1, whereas decomposition and rationale-supervision variants yield smaller, recipe-dependent changes. These findings support auditable, explicitly supervised evidence interfaces for fixed-candidate multi-hop QA and motivate blind, matched, multi-seed evaluation of the remaining small differences.}


\maketitle

\section{Introduction}\label{sec:introduction}

Question answering (QA) has progressed from span extraction over a supplied passage to knowledge-intensive settings in which a system must locate, combine, and justify information distributed across multiple sources. Pretrained encoders such as BERT established strong task adaptation through supervised fine-tuning \cite{devlin2019bert}; instruction-tuned large language models (LLMs) now additionally offer in-context learning and free-form reasoning \cite{grattafiori2024llama3}. These capabilities make LLMs attractive components of knowledge systems, but they do not eliminate the need to control what evidence reaches the model. Retrieval-augmented generation (RAG) can supply external information \cite{lewis2020rag}, yet the generator must still distinguish relevant evidence from plausible distractors.

Multi-hop QA (MHQA) makes this distinction observable. HotpotQA pairs questions with paragraph- and sentence-level supporting-fact annotations and, in its distractor setting, supplies a bounded set of gold and distractor Wikipedia paragraphs \cite{yang2018hotpotqa}. Answering often requires a bridge through an intermediate entity or a comparison between facts from different documents. A model can therefore fail because it selected the wrong passage, omitted a supporting sentence, combined evidence incorrectly, or generated an answer in the wrong form. Reporting only answer accuracy collapses these failure modes and obscures the interface between information access and reasoning.

The experiments reveal a concrete context-noise bottleneck: across Llama~3.1~8B Instruct and Llama~3.1~70B Instruct, replacing gold supporting facts with every supplied candidate paragraph reduces answer F1 by 17.32--21.43 points. Increasing reader size therefore raises the clean-evidence ceiling but does not remove sensitivity to distractors. This observation motivates an architecture that filters the benchmark-provided candidate set before answer generation. Because HotpotQA is based on general-domain Wikipedia and supplies the candidate pool, we formulate the problem as fixed-candidate evidence selection rather than domain-specific or open-corpus retrieval.

We propose \emph{Bactrainus}, a modular selector--reader system. A paragraph selector first identifies relevant passages; a sentence selector then predicts supporting facts; and an independently adapted reader generates the answer from the selected evidence. We also test two auxiliary ideas that are often conflated with the core architecture: question decomposition for evidence selection and teacher-generated chain-of-thought (CoT) rationales for reader training. We use the more precise term \emph{rationale supervision} because no teacher logits are matched and the setup is not classical knowledge distillation. In the reported comparisons, the largest gaps coincide with modularization and reader adaptation, whereas decomposition has a small effect and rationale supervision is not consistently beneficial.

The methodological contribution is system-level rather than the invention of each individual component. Paragraph selection, decomposition, rationale supervision, and LoRA are established techniques. Bactrainus places them under one task representation, makes their paragraph-, sentence-, and answer-level interfaces explicit, and studies where performance is gained or lost within a fixed candidate set.

This work makes four contributions:

\begin{itemize}
\item It quantifies the effect of progressively noisier evidence on Llama~3.1~8B Instruct and Llama~3.1~70B Instruct readers, showing that irrelevant context erodes 17.32--21.43 answer-F1 points even when the gold passages remain present.
\item It operationalizes a unified parameter-efficient paragraph-selector--question-decomposer--sentence-selector--reader framework and shows that a task-adapted instruction-tuned LLM can serve as an effective evidence retriever over a supplied candidate pool: without oracle evidence, the decomposed cascade reaches 98.24 paragraph F1 and 89.63 supporting-fact F1. This claim is specific to fixed-candidate selection and does not imply open-domain corpus-retrieval performance.
\item It conducts a systematic evaluation spanning 26 foundation models, five context conditions, direct and CoT prompting, four adapted readers, five selector configurations, and six selector--reader integration scenarios.
\item It uses matched descriptive contrasts to examine how modularization, reader scale, question decomposition, evidence granularity, and teacher-generated rationale supervision are associated with the observed differences, without treating single-run contrasts as causal estimates.
\end{itemize}

\section{Related work}\label{sec:related}

\subsection{Multi-hop QA with evidence supervision}\label{subsec:benchmarks}

Multi-hop benchmarks differ not only in reasoning depth but also in the evidence interface they expose. HotpotQA provides answer labels together with paragraph- and sentence-level supporting facts for bridge and comparison questions \cite{yang2018hotpotqa}. 2WikiMultiHopQA adds explicit reasoning paths across Wikipedia sources \cite{ho2020twiki}, whereas MuSiQue composes two- to four-hop questions while reducing common shortcut cues \cite{trivedi2022musique}. These datasets support complementary research questions, but results are not interchangeable when one setting supplies a candidate pool and another requires corpus-level retrieval.

This study uses HotpotQA's English distractor setting. Here, ``selection'' means identifying relevant paragraphs and sentences within the benchmark-provided candidate set; it does not include a first-stage search over Wikipedia. This boundary makes paragraph, supporting-fact, answer, and joint metrics directly observable and permits controlled changes to the evidence passed to a reader.

\subsection{Explicit evidence interfaces in HotpotQA}\label{subsec:retrieverreader}

HotpotQA systems commonly predict supporting facts together with the answer. HGN propagates information over paragraph, sentence, and entity nodes and jointly supervises paragraph selection, supporting-fact extraction, and answer prediction \cite{fang2020hgn}. Quark instead demonstrates a comparatively simple pipeline of sentence scoring, answer prediction, and answer-conditioned support prediction \cite{groeneveld2020quark}. FE2H separates selection and reading while organizing training from easier to harder examples \cite{li2023fe2h}, and Smoothing R3 modifies the objective and curriculum to improve generalization \cite{yin2023smoothing}.

Beam Retrieval learns multi-hop retrieval end to end and retains multiple partial passage chains during search \cite{zhang2024beam}. Bactrainus occupies a different point in this design space: it independently adapts generative Llama~3.1 selectors and readers and exposes both paragraph- and sentence-level handoffs before answer generation. The comparison therefore concerns alternative evidence interfaces, not equivalent retrieval scopes or training procedures.

\subsection{Iterative retrieval and question decomposition}\label{subsec:decomposition}

Open-domain MHQA systems often update the information need between hops. IRCoT alternates retrieval with CoT generation \cite{trivedi2023ircot}; Generate-then-Ground proposes intermediate question--answer content and then verifies it against retrieved documents \cite{shi2024gtg}; and EfficientRAG replaces repeated large-generator calls with learned iterative query generation and filtering \cite{zhuang2024efficientrag}. Question-decomposition methods instead express the information need as ordered subquestions, either through modular prompting \cite{khot2023decomposed} or decomposition-aware retrieval and reranking \cite{ammann2025qd}.

In Bactrainus, decomposition is an optional, single-pass input to supporting-sentence selection over a fixed candidate set. It does not control corpus retrieval and does not replace the original question. This placement isolates decomposition from first-stage retrieval recall and allows its contribution to be evaluated at the sentence-selection interface.

\subsection{Rationale supervision and faithfulness}\label{subsec:rationales}

CoT prompting elicits intermediate natural-language steps \cite{wei2022cot}; STaR bootstraps training from self-generated rationales that support correct answers \cite{zelikman2022star}; and Distilling Step-by-Step transfers teacher-generated labels and rationales to smaller models \cite{hsieh2023distilling}. Such targets constitute natural-language rationale supervision rather than classical logit matching.

Plausible rationales should not automatically be treated as faithful explanations of the computation that produced an answer. Controlled studies show that CoT explanations can omit influential input cues or fail to causally support the prediction \cite{turpin2023unfaithful,paul2024faithfulness}. Bactrainus therefore uses teacher-generated rationales only as an auxiliary training target and evaluates their effect on answer prediction; it does not claim that the generated text is a faithful process trace.

\subsection{Context interference in long inputs}\label{subsec:contextinterference}

The presence of relevant information does not guarantee that a long-context model will use it reliably. Controlled placement experiments show that performance can decline when relevant evidence is surrounded by additional context or appears away from favorable positions \cite{liu2024lost}. In multi-hop QA, distractors create a related but more structured interference problem because plausible paragraphs can introduce competing entities and relations. Bactrainus studies this effect while holding the gold evidence fixed and progressively adding benchmark-provided distractors, then tests whether an explicit sentence-level interface can reduce the resulting burden on the reader.

\section{Task and method}\label{sec:method}

\subsection{Task definition and data}\label{subsec:task}

Let a HotpotQA instance be $(q,D,a,S,P)$, where $q$ is a question, $D=\{d_1,\ldots,d_m\}$ is the benchmark-provided candidate set with $m\in\{2,\ldots,10\}$ in the released data, $a$ is the answer, $S$ is the set of gold supporting facts represented by paragraph title and sentence index, and $P\subset D$ is the set of gold paragraphs. Questions are either \emph{bridge} questions, which require resolving an intermediate entity, or \emph{comparison} questions, which compare attributes drawn from different sources.

We use the official English distractor split: 90,447 training instances and 7,405 development instances. The training set contains 56,814 medium, 17,972 easy, and 15,661 hard instances. The development set contains 4,990 questions with two supporting facts, 1,774 with three, and 641 with four or more. Most records contain ten candidates (89,609 training and 7,345 development instances); the implementation preserves the smaller candidate sets rather than padding them.

This is a fixed-candidate evidence-selection task. No module searches all of Wikipedia at inference time. That boundary is essential for interpreting the results and for comparing Bactrainus with open-domain systems.

\subsection{Bactrainus architecture}\label{subsec:architecture}

\begin{figure*}[t]
\centering
\includegraphics[width=0.98\textwidth]{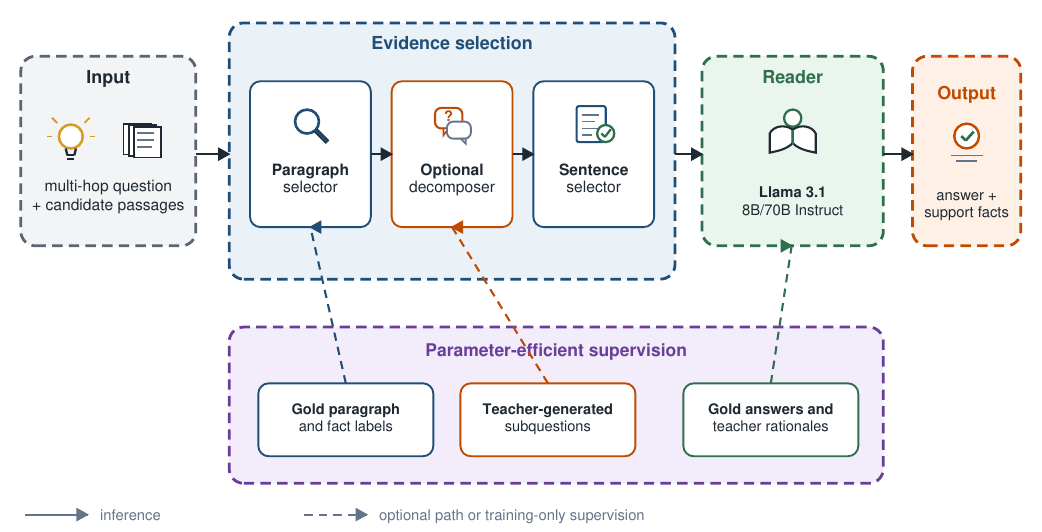}
\caption{Bactrainus architecture and training signals, adapted from the authors' original selector--reader diagram in the thesis; all icon artwork was redrawn specifically for this article using vector primitives. Colored dashed rectangles separate the input, evidence-selection modules, reader, output, and parameter-efficient supervision. Solid arrows denote inference, whereas dashed arrows denote optional paths or training-only supervision. The reader is instantiated with Llama~3.1~8B Instruct or Llama~3.1~70B Instruct.}
\label{fig:architecture}
\end{figure*}

Figure~\ref{fig:architecture} summarizes the inference path and the supervision used by each module. The selectors, decomposer, and smaller readers are initialized from \texttt{meta-llama/Meta-Llama-3.1-8B-Instruct}; the larger reader and teacher use \texttt{meta-llama/Meta-Llama-3.1-70B-Instruct} \cite{grattafiori2024llama3}. All trainable modules are adapted with low-rank adaptation (LoRA) \cite{hu2022lora}. Hereafter, ``Bactrainus 8B'' and ``Bactrainus 70B'' are explicit shorthand for configurations derived from Llama~3.1~8B Instruct and Llama~3.1~70B Instruct, respectively. Let paragraph $d_j=(t_j,s_{j1},\ldots,s_{jn_j})$ contain title $t_j$ and indexed sentences $s_{jk}$. Bactrainus implements the following maps:

\begin{align}
\hat P &= f_{P}(q,D;\theta_P), \label{eq:paragraph}\\
U &= f_{Q}(q,\hat P;\theta_Q), \label{eq:decompose}\\
\hat S &= f_{S}(q,\hat P,U;\theta_S), \label{eq:sentence}\\
\hat a &= f_{R}\!\left(q,C(\hat P,\hat S);\theta_R\right), \label{eq:reader}
\end{align}

where $\hat P$ is the selected paragraph set, $U=(u_1,\ldots,u_r)$ is an optional ordered list of subquestions, $\hat S\subseteq\{(t_j,k)\}$ is the predicted supporting-fact set, and $C$ serializes evidence for the reader. We evaluate two interfaces: $C_{\mathrm{fact}}(\hat S)$ contains only selected sentences, whereas $C_{\mathrm{para}}(\hat P)$ contains their full paragraphs. Without decomposition, $U=\varnothing$. A single-stage selector instead predicts $(\hat P,\hat S)=f_{PS}(q,D;\theta_{PS})$, and the all-in-one baseline predicts $(\hat S,\hat a)=f_J(q,D;\theta_J)$.

The modular design corresponds to the conditional factorization

\begin{equation}
\begin{split}
p(\hat P,U,\hat S,\hat a\mid q,D)
={}&p_{\theta_P}(\hat P\mid q,D)\,
p_{\theta_Q}(U\mid q,\hat P)\\
&\times p_{\theta_S}(\hat S\mid q,\hat P,U)\,
p_{\theta_R}(\hat a\mid q,C),
\end{split}
\label{eq:factorization}
\end{equation}

which exposes paragraph, sentence, and answer errors separately. The factorization is an architectural assumption rather than a claim of statistical independence: upstream predictions determine the downstream conditioning variables.

\subsubsection{Paragraph and sentence selectors}\label{subsubsec:selectors}

The paragraph selector maps $(q,D)$ to a minimal set of relevant paragraphs $\hat P$. In the single-stage variant, one model predicts both $\hat P$ and $\hat S$ directly from all supplied candidates. In the cascade, paragraph selection and sentence selection are independently supervised generations. The sentence selector receives $q$, $\hat P$, and optionally $U$, then emits exact title--sentence-index pairs. Outputs are parsed into HotpotQA's canonical supporting-fact representation; malformed or nonexistent indices count as errors.

After paragraph selection, the optional decomposer generates $U$. During target construction, teacher models generate concise subquestions from the original question and its relevant evidence; the Llama~3.1~70B Instruct teacher is used for the 15,661 hard instances and the Llama~3.1~8B Instruct teacher for the remaining training examples. At inference time, the adapted decomposer operates on the question and selected paragraphs, and its subquestions condition the sentence selector. Thus, decomposition refines fine-grained evidence identification without replacing the original question.

\subsubsection{Reader and rationale supervision}\label{subsubsec:reader}

The reader maps the question and selected evidence to a concise answer. The principal interface supplies predicted supporting sentences; a diagnostic variant supplies full selected paragraphs. Direct readers are adapted against the gold answer. Rationale-supervised readers generate the answer followed by a concise evidence-grounded rationale, matching the serialization in Appendix~\ref{app:prompts}. One Llama~3.1~8B Instruct variant uses rationales generated by a Llama~3.1~8B Instruct teacher over the full training set. A second continues the direct Llama~3.1~8B Instruct checkpoint for one epoch on 15,661 hard instances with rationales from a Llama~3.1~70B Instruct teacher. Because this continuation uses a different subset and training schedule, it is an ablation of a practical recipe rather than a perfectly controlled causal estimate of teacher size.

For module $m\in\{P,Q,S,R,PS,J\}$ with serialized input $x_i^{(m)}$ and target tokens $y_i^{(m)}$, supervised adaptation minimizes

\begin{equation}
\mathcal L_m(\theta_m)=-\sum_{i=1}^{N_m}\sum_{t=1}^{|y_i^{(m)}|}
\log p_{\theta_m}\!\left(y_{i,t}^{(m)}\mid x_i^{(m)},y_{i,<t}^{(m)}\right).
\label{eq:sft}
\end{equation}

For a direct reader, $y_i^{(R)}=[a_i]$; for a rationale-supervised reader, $y_i^{(R)}=[a_i;r_i]$, where $r_i$ is teacher-generated natural-language supervision. This is rationale supervision, not classical logit matching and not a claim that $r_i$ is a faithful mechanistic trace.

\subsection{Optimization and inference}\label{subsec:optimization}

Table~\ref{tab:readerconfig} reports the reader configurations. ``Epochs'' denotes complete passes through the respective training set. LoRA is applied to attention projections and multilayer-perceptron projections; the Llama~3.1~8B Instruct variants also adapt the language-model head. The selector configurations use the same learning-rate family and rank 64, with maximum sequence lengths of 4,096 for paragraph selection, 1,024 for sentence selection, and 2,048 for decomposition; their complete settings appear in Appendix~\ref{app:complete-results}.

For a frozen weight matrix $W_0\in\mathbb R^{d_{\mathrm{out}}\times d_{\mathrm{in}}}$, LoRA parameterizes the adapted matrix as

\begin{equation}
W=W_0+\frac{\alpha}{r}BA,
\qquad
B\in\mathbb R^{d_{\mathrm{out}}\times r},\quad
A\in\mathbb R^{r\times d_{\mathrm{in}}},
\label{eq:lora}
\end{equation}

and optimizes $A$ and $B$ while keeping $W_0$ fixed. This reduces trainable memory from $d_{\mathrm{out}}d_{\mathrm{in}}$ parameters per matrix to $r(d_{\mathrm{out}}+d_{\mathrm{in}})$.

\begin{table*}[t]
\caption{Reader adaptation configurations. Effective batch is per-device batch multiplied by gradient accumulation; distributed replication is not included. QKVO denotes the attention query, key, value, and output projections.}\label{tab:readerconfig}
\centering
\small
\resizebox{\textwidth}{!}{%
\begin{tabular}{@{}lrrrrrrrrl@{}}
\toprule
Reader & Train $n$ & Epochs & Batch & Accum. & Max len. & LR & LoRA $r$ & $\alpha$ & Targets \\
\midrule
Bactrainus (Llama 3.1 8B) & 90,447 & 2 & 8 & 32 & 512 & $10^{-4}$ & 64 & 128 & QKVO, MLP, head \\
Llama 3.1 8B + 8B-teacher rationale & 90,447 & 2 & 4 & 16 & 1,024 & $10^{-4}$ & 64 & 128 & QKVO, MLP, head \\
Llama 3.1 8B + 70B-teacher rationale & 15,661 & 1 & 4 & 16 & 1,024 & $10^{-4}$ & 64 & 32 & QKVO, MLP, head \\
Bactrainus (Llama 3.1 70B) & 90,447 & 1 & 1 & 8 & 512 & $10^{-4}$ & 16 & 16 & QKVO, MLP \\
\bottomrule
\end{tabular}
}
\end{table*}

Training used cosine learning-rate decay, LoRA dropout 0.05, and warm-up ratio 0.03, except the Llama~3.1~70B-rationale continuation, which used 0.10. Experiments ran on a server equipped with four NVIDIA A100 80-GB GPUs. Most Llama~3.1~8B jobs used one device; model-parallel large-model jobs used up to three devices.

Generation was near deterministic but varied slightly by backend. Prompted/API experiments used temperature 0.01 and top-$p$ 0.99. Fine-tuned selector and reader pipelines used temperature 0.0001 and top-$p$ 0.90, while the integrated one-step baseline used 0.0001 and 0.99.

Hard distractors for the context ablation are selected by cosine similarity. With embedding function $e(\cdot)$,

\begin{equation}
\begin{aligned}
s(q,d_j)&=\frac{e(q)^\top e(d_j)}{\lVert e(q)\rVert_2\lVert e(d_j)\rVert_2},\\
K_i&=\min\!\left(2,|D_i\setminus P_i|\right),\\
D_{\mathrm{hard},i}&=\operatorname*{TopK_i}_{d_j\in D_i\setminus P_i}\,s(q_i,d_j).
\end{aligned}
\label{eq:harddistractor}
\end{equation}

\subsection{Foundation-model screening and proprietary-score calibration}\label{subsec:snapshot}

Because proprietary-model cost limited direct evaluation to a fixed subset $H_{700}\subset H_{\mathrm{dev}}$ of 700 questions, their reported values are calibrated full-development-equivalent estimates. Let $\mathcal O$ be the set of open-source reference models evaluated on both $H_{700}$ and all $H_{\mathrm{dev}}$. For metric $M\in\{\mathrm{EM},\mathrm{F1}\}$, we compute

\begin{equation}
\kappa_M=
\frac{|\mathcal O|^{-1}\sum_{o\in\mathcal O}M_o(H_{\mathrm{dev}})}
{|\mathcal O|^{-1}\sum_{o\in\mathcal O}M_o(H_{700})},
\qquad
\widetilde M_c=\kappa_M M_c(H_{700}),
\label{eq:closedcalibration}
\end{equation}

for each closed-source model $c$. The ratio is deliberately full-set over 700-subset, and EM and F1 are calibrated separately. These values support broad model screening only; all subsequent architecture claims use directly evaluated Llama~3.1 experiments.

This estimator assumes that the pooled full-set/subset ratio observed for the open-model reference pool transfers to each proprietary model. Appendix~\ref{app:h700precision} gives a finite-population precision calculation, proves exact recovery under the common multiplicative-shift model, and separates item-sampling from coefficient uncertainty. Model-specific transport remains an assumption, so the resulting values are coarse screening indicators rather than a basis for significance claims or fine-grained ranking of systems with close scores.

\subsection{Evaluation measures}\label{subsec:metrics}

We use the official HotpotQA normalization: lowercase text, remove punctuation and English articles, and collapse whitespace. Answer exact match (EM) is

\begin{equation}
\mathrm{EM}(\hat a,a)=\mathbb{1}[N(\hat a)=N(a)],
\end{equation}

where $N$ is the normalization function. With normalized answer-token multisets $T(\hat a)$ and $T(a)$ and overlap count $c_A$, answer precision, recall, and F1 are

\begin{equation}
P_A=\frac{c_A}{|T(\hat a)|},\qquad
R_A=\frac{c_A}{|T(a)|},\qquad
F_{1,A}=\frac{2P_AR_A}{P_A+R_A}.
\label{eq:answerf1}
\end{equation}

For supporting facts, $P_S=|\hat S\cap S|/|\hat S|$ and $R_S=|\hat S\cap S|/|S|$, with the same harmonic mean; supporting-fact EM is $\mathbb 1[\hat S=S]$. The official joint measures are

\begin{equation}
\mathrm{EM}_{J}=\mathrm{EM}_{A}\mathrm{EM}_{S},\quad
P_J=P_AP_S,\quad R_J=R_AR_S,\quad
F_{1,J}=\frac{2P_JR_J}{P_J+R_J}.
\label{eq:joint}
\end{equation}

Zero denominators follow the official evaluator's conventions.

All values are percentages and all differences are percentage points. For the Llama~3.1~8B Instruct reader-adaptation comparison, we additionally use a paired nonparametric percentile bootstrap with 4,000 resamples and seed 20,260,807 over 7,402 paired instances. This post-hoc interval is reported only for that comparison and is not generalized to selector or joint results.

\subsection{Use of generative AI in manuscript preparation}\label{subsec:genai}

OpenAI Codex was used during manuscript writing and analysis to assist with language editing, internal-consistency checks, reproducibility review, and code cleanup. It was not used to generate experimental observations or reported numerical results. The authors reviewed and edited all assisted material, verified every reported result against their research records, and take full responsibility for the final manuscript.

\section{Experimental design}\label{sec:design}

The experiments address four questions. RQ1 asks which model families provide strong readers and how evidence quality and distractor volume affect readers of different sizes. RQ2 asks whether prompting or parameter-efficient task adaptation best improves a controlled reader. RQ3 asks how single-stage selection, cascaded selection, and decomposition affect evidence prediction. RQ4 asks which component drives integrated performance: modularization, evidence granularity, decomposition, rationale supervision, or reader scale.

All reported evaluation scores are obtained on the official HotpotQA distractor development set; no blind test-set result is claimed. The same development set was inspected while comparing model families, prompting choices, module variants, and integrated configurations. Consequently, the displayed maximum is a post-selection development-set estimate rather than an unbiased estimate of final-test generalization. We interpret these experiments as an internal benchmark analysis and reserve external-generalization claims for future held-out evaluation.

\subsection{Context and prompting ablations}\label{subsec:contextdesign}

For RQ1, each reader receives one of five inputs: the question alone; gold supporting facts; the gold paragraphs; the gold paragraphs plus two difficult distractors; or all supplied candidate paragraphs. Difficult distractors are the non-gold paragraphs with highest question similarity under \texttt{gte-large-en-v1.5}, a general text-embedding model \cite{li2023gte,alibaba2024gtelargev15}. The comparison therefore holds the question and answer fixed while increasing irrelevant content.

We also test zero-, one-, two-, four-, and eight-shot prompting with and without a CoT instruction. The eight fixed demonstrations contain two hard, five medium, and one easy question and cover both bridge and comparison types; they are not sampled exclusively from the hard split. Prefixes are nested except that the recorded eight-shot order is $[0,1,2,4,5,6,7,3]$. This detail is included because demonstration identity can be more influential than demonstration count.

\subsection{Module and pipeline ablations}\label{subsec:pipelinedesign}

Selector experiments compare: (i) a single-stage selector, (ii) an independent paragraph selector, (iii) a sentence selector with oracle paragraphs, (iv) the predicted two-stage cascade, and (v) the cascade augmented with generated subquestions. Reader experiments assume gold supporting facts so that adaptation can be evaluated separately from selector errors.

End-to-end experiments combine the modules in six scenarios:

\begin{enumerate}
\item an all-in-one Llama~3.1~8B Instruct model predicts evidence and answer jointly;
\item a single-stage selector passes supporting facts to a reader;
\item the paragraph--sentence cascade passes supporting facts to a reader;
\item the cascade passes full selected paragraphs to a reader;
\item the decomposed cascade passes supporting facts to a reader; and
\item the decomposed cascade passes full selected paragraphs to a reader.
\end{enumerate}

For each modular scenario, the reader is varied among direct Llama~3.1~8B Instruct, the same Llama~3.1~8B reader with Llama~3.1~70B-teacher rationale supervision, and direct Llama~3.1~70B Instruct. The main external comparison is Beam Retrieval \cite{zhang2024beam}. Because published systems can differ in preprocessing, checkpoints, and dev-versus-test evaluation, we call this a comparison with reported HotpotQA distractor scores rather than a current leaderboard or universal state-of-the-art claim.

\begin{figure*}[t]
\centering
\includegraphics[width=0.98\textwidth]{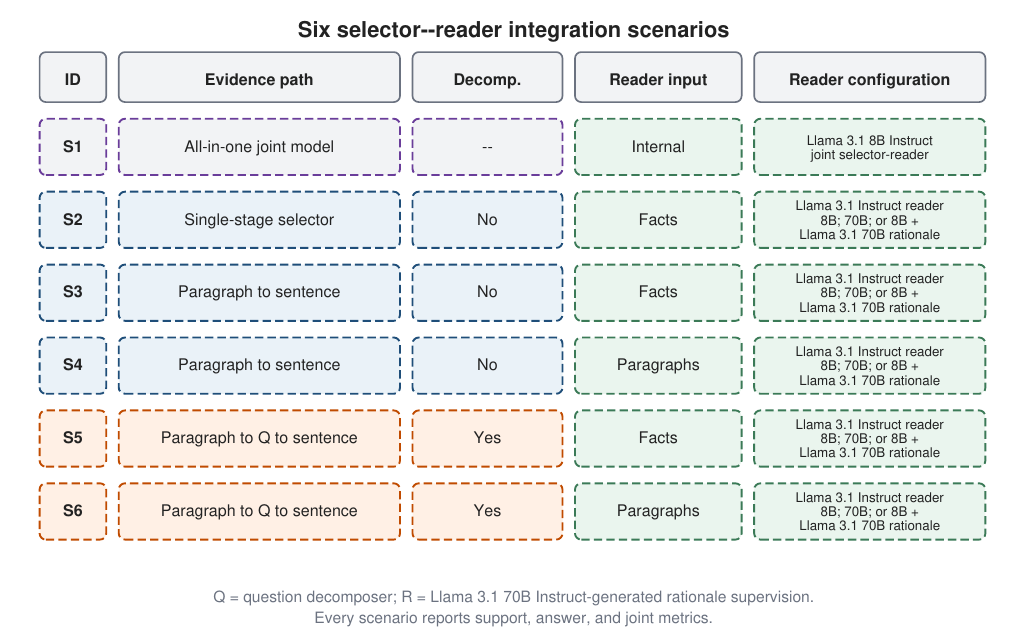}
\caption{Six integration scenarios, separated into colored components with dashed boundaries. Scenarios 2--6 pair each evidence path with direct Llama~3.1~8B Instruct, rationale-supervised Llama~3.1~8B Instruct, and direct Llama~3.1~70B Instruct readers. ``Facts'' and ``paragraphs'' denote the two reader interfaces $C_{\mathrm{fact}}$ and $C_{\mathrm{para}}$.}
\label{fig:scenarios}
\end{figure*}

\section{Results}\label{sec:results}

\subsection{Foundation-model screening motivates the reader scale anchors}\label{subsec:screening}

The complete 26-model screening appears in Table~\ref{tab:allmodels}. Among the calibrated proprietary estimates, GPT-4o obtains 67.54 EM/83.44 F1. Among open models below 10B parameters, Llama~3.1~8B Instruct reaches 60.11/74.52; around 70B, Llama~3.1~70B Instruct reaches 65.60/80.04. Llama~3.1~405B Instruct reaches 67.46/82.56 but has a substantially larger parameter footprint; serving cost was not measured, and it was not retained for the controlled Llama~3.1~8B/70B study. The two selected checkpoints therefore share a model family while representing distinct capacity regimes.

The proprietary scores are screening estimates derived from the 700-question API subset through Eq.~\eqref{eq:closedcalibration}; they are not used as baselines for the selector--reader gain. The screening documents reader selection for the controlled experiments rather than a ranking of mutable proprietary services. The model-version cutoff is reported once in the note to Table~\ref{tab:allmodels}.

\subsection{Context noise is the dominant reader bottleneck}\label{subsec:contextresults}

\begin{figure*}[t]
\centering
\includegraphics[width=0.98\textwidth]{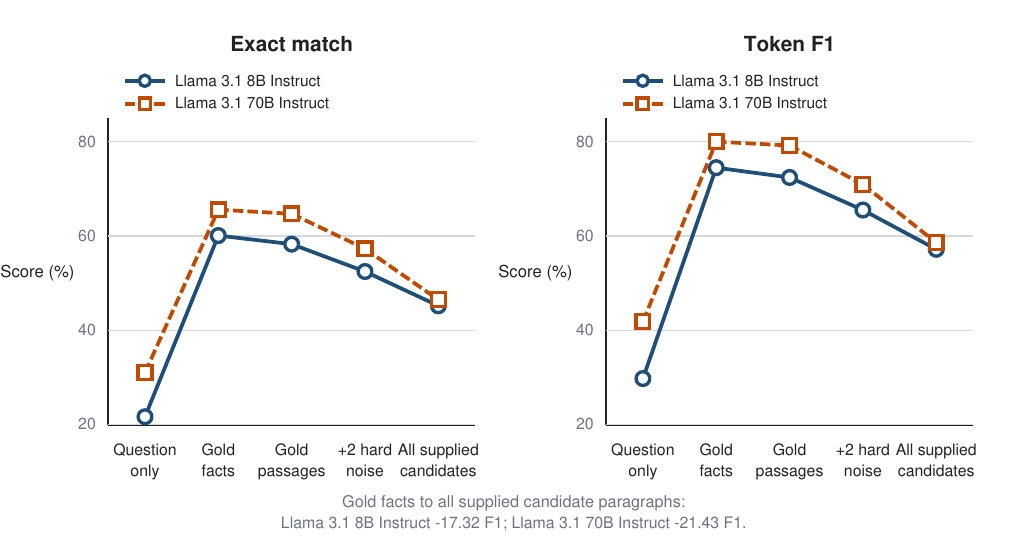}
\caption{Exact match and token F1 as evidence becomes noisier. Both readers peak with gold supporting facts and degrade as irrelevant text is added. From gold facts to all supplied candidates, Llama~3.1~8B Instruct loses 14.90 EM/17.32 F1 and Llama~3.1~70B Instruct loses 19.10 EM/21.43 F1.}
\label{fig:context}
\end{figure*}

Table~\ref{tab:context} and Fig.~\ref{fig:context} answer RQ1. With only the question, neither model has enough factual context. Gold supporting sentences yield the strongest scores. Supplying the full gold paragraphs already lowers Llama~3.1~8B Instruct F1 by 2.08 points and Llama~3.1~70B Instruct F1 by 0.81 points, even though no paragraph-level distractor is present. Adding two similarity-selected distractors causes a further loss, and using all supplied candidate paragraphs reduces F1 to 57.20 and 58.61.

\begin{table*}[t]
\caption{Answer EM/F1 under controlled evidence conditions on HotpotQA distractor dev. The strongest value in each model row is bold.}\label{tab:context}
\centering
\small
\resizebox{\textwidth}{!}{%
\begin{tabular}{@{}lccccc@{}}
\toprule
Model & Question only & Gold support facts & Gold paragraphs & Gold + 2 hard distractors & All candidates \\
\midrule
Llama 3.1 8B Instruct & 21.66/29.76 & \textbf{60.11/74.52} & 58.29/72.44 & 52.48/65.53 & 45.21/57.20 \\
Llama 3.1 70B Instruct & 31.02/41.85 & \textbf{65.60/80.04} & 64.74/79.23 & 57.31/70.96 & 46.50/58.61 \\
\bottomrule
\end{tabular}
}
\end{table*}

The larger model is consistently stronger except under the noisiest F1 condition, where the gap narrows to 1.41 points. Scaling therefore improves the clean-evidence ceiling but does not confer proportional robustness to distractors. This observation supports filtering evidence before reasoning instead of relying on reader capacity alone.

Define the absolute noise loss and retained-performance ratio for reader $m$ as

\begin{equation}
\Delta_{\mathrm{noise}}^{(m)}=F_{1,m}^{\mathrm{gold\ facts}}-F_{1,m}^{\mathrm{all\ candidates}},
\qquad
\rho_{\mathrm{noise}}^{(m)}=\frac{F_{1,m}^{\mathrm{all\ candidates}}}{F_{1,m}^{\mathrm{gold\ facts}}}.
\label{eq:noiserobustness}
\end{equation}

For Llama~3.1~8B Instruct, $(\Delta_{\mathrm{noise}},\rho_{\mathrm{noise}})=(17.32,0.768)$; for Llama~3.1~70B Instruct, it is $(21.43,0.732)$. The larger reader has the higher clean-evidence ceiling but retains a smaller fraction of that ceiling under the full distractor context.

The stored Llama~3.1~8B Instruct predictions can be aligned row-wise with the official question-type metadata. Table~\ref{tab:questiontype} shows that both bridge and comparison questions exhibit substantial context-noise loss and adaptation gain. The bootstrap intervals are paired within each type, and the analysis is post hoc; it is intended to test whether the aggregate pattern is confined to one major question category, not to define a new test set.

\begin{table*}[t]
\caption{Question-type analysis for the Llama~3.1~8B Instruct reader. Values are answer F1; brackets are paired 95\% percentile-bootstrap intervals for the displayed difference (4,000 resamples; base seed 20,260,808). Rows with missing generations are excluded pairwise.}\label{tab:questiontype}
\centering
\small
\resizebox{\textwidth}{!}{%
\begin{tabular}{@{}lrrrrrr@{}}
\toprule
Type & $n_{\mathrm{adapt}}/n_{\mathrm{noise}}$ & Zero-shot, gold facts & Adapted, gold facts & Adaptation $\Delta$ [95\% CI] & All candidates & Noise loss $\Delta$ [95\% CI] \\
\midrule
Bridge & 5,915/5,912 & 74.32 & 86.19 & $+11.87$ [10.98, 12.75] & 56.09 & $-18.23$ [$-19.31$, $-17.14$] \\
Comparison & 1,487/1,487 & 75.30 & 87.61 & $+12.31$ [10.41, 14.18] & 61.74 & $-13.56$ [$-15.66$, $-11.46$] \\
\bottomrule
\end{tabular}}
\end{table*}

Performance is also not monotonic in the number of gold facts. For example, the Llama~3.1~70B Instruct reader scores 79.92 F1 on two-fact questions, 79.20 on three-fact questions, and 83.28 on questions with four or more facts. Table~\ref{tab:factcount} reports all five models and bins. Fact count is entangled with question type and answer form and should not be interpreted as an isolated measure of reasoning depth.

\subsection{Few-shot examples help, whereas CoT prompting is unstable}\label{subsec:promptresults}

\begin{figure*}[t]
\centering
\includegraphics[width=0.96\textwidth]{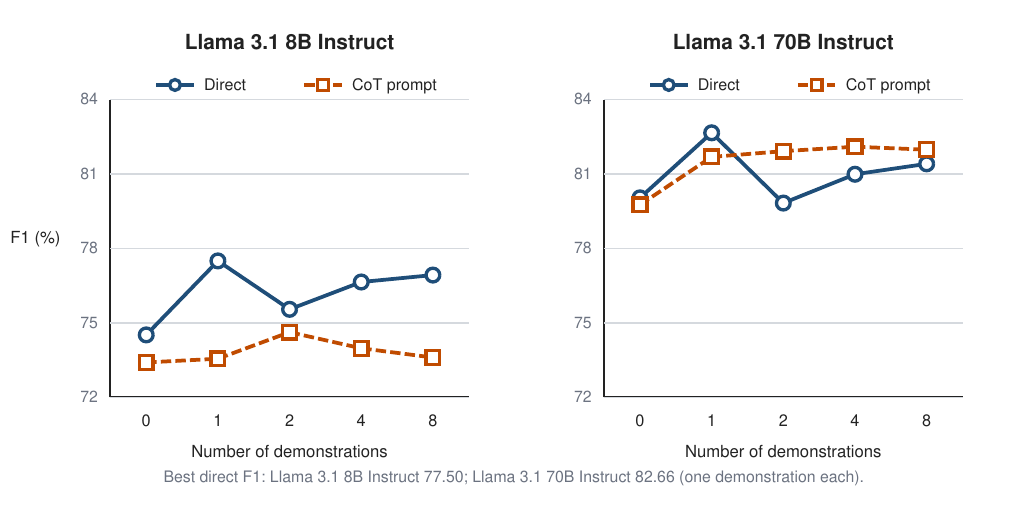}
\caption{Answer F1 versus demonstration count with direct and CoT prompts. CoT reduces performance for Llama~3.1~8B Instruct at every count. For Llama~3.1~70B Instruct, it is more stable after two demonstrations but never exceeds the best one-shot direct prompt. Exact EM/F1 values are reported in Table~\ref{tab:promptfull}.}
\label{fig:prompting}
\end{figure*}

Figure~\ref{fig:prompting} and Table~\ref{tab:promptfull} report the complete direct and CoT prompting study. Without CoT, one demonstration is best for both sizes: it improves Llama~3.1~8B Instruct from 74.52 to 77.50 F1 and Llama~3.1~70B Instruct from 80.04 to 82.66. Additional demonstrations do not produce monotonic gains.

CoT prompting lowers every Llama~3.1~8B Instruct setting, with the largest gap at one shot (3.94 F1 points). For Llama~3.1~70B Instruct, CoT improves over the corresponding direct prompt at two, four, and eight demonstrations but remains 0.56 points below the best direct result. The finding is not that reasoning traces are universally ineffective; rather, generic CoT instructions do not reliably compensate for evidence-selection errors and may consume context or induce verbose, harder-to-parse outputs in a smaller model.

\subsection{Reader adaptation yields large oracle-evidence gains}\label{subsec:readerresults}

Table~\ref{tab:readers} answers RQ2 under gold supporting facts. Direct adaptation raises the Llama~3.1~8B Instruct reader by 13.91 EM and 11.94 F1 over zero-shot prompting. On 7,402 paired instances, the bootstrap estimate is 13.94 EM (95\% CI 12.90--15.01) and 11.96 F1 (95\% CI 11.17--12.78), consistent with the aggregate effect. The direct Llama~3.1~70B Instruct reader reaches 90.01 F1, 9.97 points above its zero-shot counterpart.

\begin{table}[t]
\caption{Reader performance with gold supporting facts. Rationale denotes teacher-generated natural-language supervision, not logit distillation.}\label{tab:readers}
\centering
\small
\begin{tabular}{@{}lcc@{}}
\toprule
Reader & Answer EM & Answer F1 \\
\midrule
Llama 3.1 8B Instruct, zero-shot & 60.11 & 74.52 \\
Bactrainus (Llama 3.1 8B) & 74.02 & 86.46 \\
Llama 3.1 8B + 8B-teacher rationale & 72.97 & 85.62 \\
Llama 3.1 8B + 70B-teacher rationale & 74.19 & 86.91 \\
Llama 3.1 70B Instruct, zero-shot & 65.60 & 80.04 \\
Bactrainus (Llama 3.1 70B) & \textbf{75.73} & \textbf{90.01} \\
\bottomrule
\end{tabular}
\end{table}

The observed rationale results vary across training recipes. Rationales from the Llama~3.1~8B teacher lower the direct Llama~3.1~8B reader by 1.05 EM/0.84 F1, whereas continuing the reader on hard examples with Llama~3.1~70B-generated rationales changes it by $+0.17$ EM/$+0.45$ F1. Because teacher size, training subset, and schedule change jointly, the comparison does not identify an independent teacher-quality or rationale effect. Direct answer supervision remains the strongest measured answer-only baseline.

\subsection{Selection is accurate at paragraph level but bottlenecked at sentence level}\label{subsec:selectorresults}

Table~\ref{tab:selectors} answers RQ3. Paragraph selection exceeds 98 F1, but supporting-fact EM remains near 66 because one missing or extra sentence makes the entire set incorrect. The oracle-paragraph sentence selector obtains 89.93 F1, 0.72 points above the predicted cascade, quantifying modest paragraph-to-sentence error propagation.

\begin{table*}[t]
\caption{Evidence-selector results on HotpotQA distractor dev. A dash means the configuration does not independently emit that level of prediction.}\label{tab:selectors}
\centering
\small
\resizebox{\textwidth}{!}{%
\begin{tabular}{@{}lcccc@{}}
\toprule
Method & Paragraph EM & Paragraph F1 & Support-fact EM & Support-fact F1 \\
\midrule
Single-stage selector & \textbf{96.83} & \textbf{98.37} & 65.74 & 89.27 \\
Paragraph selector & 96.64 & 98.24 & -- & -- \\
Sentence selector, oracle paragraphs & -- & -- & \textbf{66.01} & \textbf{89.93} \\
Predicted two-stage cascade & 96.64 & 98.24 & 65.55 & 89.21 \\
Two-stage + decomposition & 96.64 & 98.24 & 65.93 & 89.63 \\
\bottomrule
\end{tabular}
}
\end{table*}

Decomposition adds 0.38 EM/0.42 F1 over the plain cascade and 0.19 EM/0.36 F1 over the single-stage selector. Without repeated seeds or per-example selector outputs suitable for paired inference, this difference should be characterized as small rather than ``notable.'' Its main value is qualitative: generated subquestions can recover some sentence-level evidence, but they do not change paragraph selection and are not the primary source of system gains.

\subsection{The largest end-to-end differences align with modularization and reader size}\label{subsec:e2e}

Table~\ref{tab:pipelines} summarizes the most informative integrated configurations. The all-in-one Llama~3.1~8B Instruct model obtains 83.31 answer F1 and 75.96 joint F1. Passing selected supporting facts to a separately adapted Llama~3.1~8B Instruct reader raises these values to 85.41 and 77.90. Replacing that reader with Llama~3.1~70B Instruct is associated with most of the remaining difference. In contrast, adding decomposition to the Llama~3.1~70B cascade changes answer F1 by 0.21 and joint F1 by 0.17.

\begin{table*}[t]
\caption{Selected end-to-end configurations. Values are EM/F1. ``Facts'' and ``paragraphs'' identify the selector--reader interface. Boldface marks the numerical maximum among reported development-set runs; it does not denote a preregistered or statistically separated winner.}\label{tab:pipelines}
\centering
\small
\resizebox{\textwidth}{!}{%
\begin{tabular}{@{}lllllll@{}}
\toprule
Scenario & Selector & Reader input & Reader & Support facts & Answer & Joint \\
\midrule
All-in-one & joint generation & internal & Llama 3.1 8B & 63.42/88.50 & 71.24/83.31 & 47.93/75.96 \\
Single-stage & direct & facts & Llama 3.1 8B & 65.74/89.27 & 73.24/85.41 & 50.84/77.90 \\
Two-stage & cascade & facts & Llama 3.1 8B & 65.55/89.21 & 73.20/85.38 & 50.73/77.86 \\
Two-stage & cascade & paragraphs & Llama 3.1 70B & 65.55/89.21 & 74.04/87.85 & 51.02/77.03 \\
Decomposed & cascade + subquestions & facts & Llama 3.1 8B & 65.93/89.63 & 73.34/85.56 & 50.88/77.99 \\
Decomposed & cascade + subquestions & facts & Llama 3.1 8B + rationale & 65.93/89.63 & 73.36/85.60 & 50.91/78.02 \\
Decomposed & cascade + subquestions & facts & Llama 3.1 70B & \textbf{65.93/89.63} & \textbf{75.07/89.01} & \textbf{51.73/79.70} \\
\bottomrule
\end{tabular}
}
\end{table*}

Passing full paragraphs to the reader is consistently weaker than passing selected facts. For the decomposed system with a Llama~3.1~70B Instruct reader, the paragraph interface yields 87.85 answer F1 and 77.03 joint F1, compared with 89.01 and 79.70 for the fact interface. This 2.67-point joint-F1 gap aligns with the context ablation and suggests both avoidable noise and a train--test interface mismatch.

The highest-scoring reported system differs from the all-in-one baseline by $+2.51/+1.13$ supporting-fact EM/F1, $+3.83/+5.70$ answer EM/F1, and $+3.80/+3.74$ joint EM/F1. Relative to the single-stage selector plus the same Llama~3.1~70B Instruct reader, however, the decomposed cascade changes supporting facts by $+0.19/+0.36$, answers by $+0.11/+0.18$, and joint scores by $+0.12/+0.14$ EM/F1. These descriptive contrasts associate the larger differences with a separated selection--reading interface and reader capacity rather than with extra stages alone.

To summarize the observed development-set gaps without causally allocating them, let $L_8$ and $L_{70}$ denote the Llama~3.1~8B Instruct and Llama~3.1~70B Instruct readers and define three descriptive joint-F1 contrasts:

\begin{align}
\Delta_{\mathrm{mod}} &= F_{1,J}(S_2,L_8)-F_{1,J}(S_1)=1.94,\\
\Delta_{\mathrm{scale}} &= F_{1,J}(S_2,L_{70})-F_{1,J}(S_2,L_8)=1.66,\\
\Delta_{Q} &= F_{1,J}(S_5,L_{70})-F_{1,J}(S_3,L_{70})=0.17.
\end{align}

The modularization contrast is numerically an order of magnitude larger than the matched decomposition contrast, while scale supplies a second sizable difference. These quantities do not identify unique causal effects: the all-in-one and modular systems differ in output factorization, supervision, and number of model invocations; the scale contrast changes capacity; and configurations were not repeated over training seeds. The $\Delta$ terms therefore summarize observed gaps rather than additive treatment effects.

\begin{figure*}[t]
\centering
\includegraphics[width=0.96\textwidth]{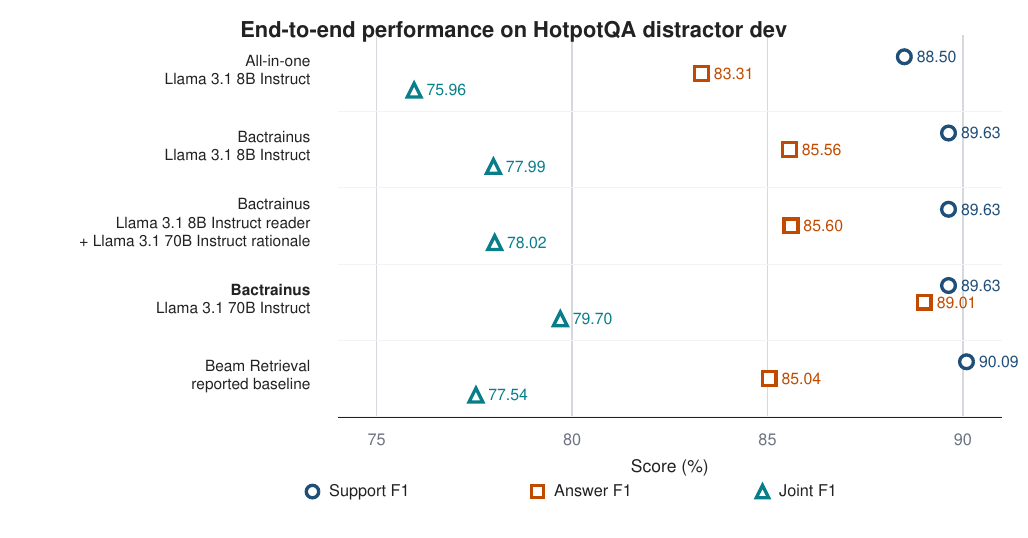}
\caption{Reported end-to-end F1 trade-off on HotpotQA distractor dev. Numerical maxima differ by metric: Bactrainus with the Llama~3.1~70B Instruct reader is highest on answer and joint F1 among displayed systems, whereas Beam Retrieval is highest on supporting-fact F1. Bold/position denotes only a numerical maximum, not protocol equivalence or statistical separation.}
\label{fig:e2e}
\end{figure*}

The numerical differences between the separately reported Bactrainus and Beam Retrieval results are $+2.38/+3.97$ answer EM/F1, $+1.20/+2.16$ joint EM/F1, and $-0.32/-0.46$ supporting-fact EM/F1 for Bactrainus. They are descriptive and do not isolate a method effect. Recent iterative RAG systems such as Generate-then-Ground and EfficientRAG are discussed conceptually rather than inserted into the numeric table because no matched evaluation under this fixed-candidate protocol was available \cite{shi2024gtg,zhuang2024efficientrag}.

\section{Discussion}\label{sec:discussion}

\subsection{What patterns accompany the gain?}\label{subsec:gain}

Three observations form a consistent diagnostic picture. First, both model sizes are highly sensitive to irrelevant context. Second, a separately adapted reader has much higher reported scores than its prompted counterpart when evidence is controlled. Third, decomposition and rationale-supervision differences are below one point. Together, these observations are consistent with the value of a narrow, supervised evidence interface, although the single-run design does not establish a unique causal allocation.

This interpretation reframes the system as a knowledge-processing pipeline rather than a generic demonstration that LLMs ``reason.'' Paragraph F1 above 98 shows that coarse relevance is comparatively easy within the supplied candidate set. Sentence-level set prediction and answer generation are harder, and their errors interact. A modular interface lets each error source be measured and optimized, which is valuable even when an end-to-end model could in principle represent the same function.

\subsection{Implications for decomposition and rationales}\label{subsec:implications}

Question decomposition has a plausible role when the original question underspecifies intermediate entities, but its small average gain suggests that subquestions are neither uniformly necessary nor uniformly reliable. A future selector could conditionally invoke decomposition based on uncertainty rather than generating subquestions for every instance. It could also score the consistency between subquestions and the original question before using them.

Rationale supervision poses a different trade-off. The two reported recipes move performance in opposite directions, but teacher size, training subset, and schedule change jointly. Future work should compare answer-only and rationale targets on identical examples, evaluate rationale faithfulness against evidence, and report the additional token and compute cost. These controls are necessary before attributing the difference to process supervision.

\subsection{Implications for practical knowledge systems}\label{subsec:practical}

The modular scenarios have a transparent structural cost. If one invocation is counted for each sequentially executed generative module, then

\begin{equation}
N_{\mathrm{calls}}(S_1,\ldots,S_6)=(1,2,3,3,4,4).
\label{eq:callcount}
\end{equation}

This invocation count is a structural proxy, not a latency measurement: token lengths, batching, parallel execution, hardware, and backend overhead differ. The modular Llama~3.1~8B Instruct run recovers 1.94 of the 3.74 joint-F1 difference between the all-in-one baseline and the highest-scoring reported configuration, making it a candidate for cost-sensitive deployment; that choice still requires direct latency and resource measurements. Because Bactrainus assumes a provided candidate set, deployment also requires an upstream corpus retriever whose recall and freshness constraints are not measured here.

\section{Limitations and threats to validity}\label{sec:limitations}

\textbf{Dataset and task scope.} Evaluation is limited to English HotpotQA distractor. The system has not been tested on 2WikiMultiHopQA, MuSiQue, domain-specific corpora, or open-domain corpus retrieval. It receives the benchmark-provided set of 2--10 candidate paragraphs, so results do not measure first-stage retrieval recall.

\textbf{Development-set reuse.} The official development set serves both as the configuration-comparison set and the final reported evaluation set. Repeated inspection may favor the selected configuration, so the maximum score should not be interpreted as blind-test performance. The question-type bootstrap is likewise post hoc and does not remove this selection limitation.

\textbf{Model endpoints and calibration.} Proprietary endpoints and mutable checkpoint aliases can change over time; the evaluation cutoff is recorded once in Table~\ref{tab:allmodels}. Closed-source values are calibrated from 700 direct evaluations via Eq.~\eqref{eq:closedcalibration}. Appendix~\ref{app:h700precision} quantifies the conditional finite-sample resolution and coefficient variance, but transfer of a pooled open-model ratio to each proprietary model remains an unverified transportability assumption. None of the architectural conclusions depends on ranking a proprietary model.

\textbf{Model and compute scope.} The controlled architecture experiments focus on Llama~3.1~8B Instruct and Llama~3.1~70B Instruct. The larger reader carries substantial memory and latency costs, yet the study does not report end-to-end latency, energy, throughput, or dollar cost. This prevents a complete efficiency comparison with iterative RAG systems or smaller encoders. Fine-tuning and RAG may also behave differently when knowledge is novel rather than recoverable from the training distribution \cite{yang2026finetune}.

\textbf{Statistical uncertainty.} Most configurations are single runs. The paired bootstrap covers the principal Llama~3.1~8B Instruct reader-adaptation contrast, but selector and joint differences below one point remain unresolved without repeated seeds. Accordingly, decomposition and rationale changes are described as small or conditional rather than definitive.

\textbf{Supervision and rationale faithfulness.} Teacher-generated subquestions and rationales are derived from labeled training examples, while inference uses model predictions. Rationale faithfulness is not independently verified. These factors limit causal claims about intrinsic reasoning ability and motivate direct process-level evaluation in future work.

\textbf{External comparisons.} Beam Retrieval and Bactrainus use different implementations and possibly different checkpoint-selection procedures. Their reported HotpotQA scores are useful reference points but are not a substitute for rerunning both systems under one codebase. Accordingly, we claim a metric trade-off among the compared results, not universal state of the art.

\section{Conclusion}\label{sec:conclusion}

Bactrainus treats multi-hop QA as a sequence of auditable knowledge-processing decisions: select passages, identify supporting sentences, and answer from compact evidence. On the HotpotQA distractor development set, the modular configurations have higher reported scores than the all-in-one Llama~3.1~8B Instruct run and counteract severe distractor-context degradation. The highest-scoring configuration reaches 89.01 answer F1 and 79.70 joint F1, while the separately reported Beam Retrieval result remains higher on supporting-fact F1. Within the reported single-run comparisons, the largest gaps coincide with modularization and reader adaptation/scale; decomposition is marginal and rationale-supervision results are recipe-dependent. A blind, matched, multi-seed evaluation with retrieval and efficiency measurements is required to test generalization beyond this internal benchmark study.

\section*{Declarations}

\subsection*{Funding}
The authors received no specific funding for this work.

\subsection*{Competing interests}
The authors declare that they have no competing interests.

\subsection*{Ethics approval and consent to participate}
Not applicable. This study uses a publicly available text benchmark and does not involve human participants, personal data collection, or animal research.

\subsection*{Consent for publication}
Not applicable.

\subsection*{Data availability}
HotpotQA is publicly available from \url{https://huggingface.co/datasets/hotpotqa/hotpot_qa} and the benchmark website \url{https://hotpotqa.github.io/} \cite{yang2018hotpotqa}. The reported experiments use the official English distractor training and development splits. The cleaned canonical Bactrainus training release contains eight ID-aligned configurations for structured records and deterministic selector/reader SFT. It is available at \href{https://huggingface.co/datasets/bactrianus/bactrainus-hotpotqa/tree/f5a70de8c280eb0cd43e21ea246d71197141c65e}{\texttt{bactrianus/\allowbreak bactrainus-hotpotqa}} (version \texttt{v1.0.0}, immutable revision \texttt{f5a70de8c280eb0cd43e21ea246d71197141c65e}). A separate source-linked release contains four teacher-supervision configurations: Llama~3.1 8B and 70B rationale traces and question decompositions. It is available at \href{https://huggingface.co/datasets/bactrianus/bactrainus-hotpotqa-teacher-traces/tree/4bc0f972978d520657f974496528667e2bb5ff50}{\texttt{bactrianus/\allowbreak bactrainus-hotpotqa-\allowbreak teacher-traces}} (version \texttt{v1.0.0}, immutable revision \texttt{4bc0f972978d520657f974496528667e2bb5ff50}). Each released teacher view retains one resolved output per source ID and records the available artifact provenance; immutable historical serving revisions were not retained and are marked accordingly. Evaluation predictions and result files are not included in either dataset release.

\subsection*{Materials availability}
Public Bactrainus model artifacts and model cards are organized in the \href{https://huggingface.co/bactrianus}{Bactrainus Hugging Face collection}. Each card identifies its own base checkpoint, intended scope, license, and access terms; the exact Llama~3.1 base-model identifiers used in this study are reported in Sect.~\ref{subsec:architecture}.

\subsection*{Code availability}
The cleaned data-construction, training, inference, pipeline, and evaluation code is public in the \href{https://github.com/Iman998/bactrainus/tree/e9e8139e958124f17cdc0dac6df65d1fc675193a}{Bactrainus GitHub repository} at immutable commit \texttt{e9e8139e958124f17cdc0dac6df65d1fc675193a}. The repository includes machine-readable configurations and a versioned prompt catalog; the complete experimental message templates are also reproduced in Appendix~\ref{app:prompts}.

\subsection*{Author contributions}
Iman Barati: conceptualization, methodology, software, investigation, data curation, visualization, and writing---original draft. Arash Ghafouri: validation and writing---review and editing. Behrouz Minaei-Bidgoli: supervision and writing---review and editing. All authors reviewed and approved the manuscript.

\begin{appendices}
\renewcommand{\theHtable}{app.\Alph{section}.\arabic{table}}
\renewcommand{\theHfigure}{app.\Alph{section}.\arabic{figure}}
\section{Complete experimental results}\label{app:complete-results}

This appendix reports every experimental family used in the study. Unless stated otherwise, scores are percentages on the English HotpotQA distractor development set and cells report EM/F1. Closed-source systems were run on the fixed 700-question subset and calibrated with Eq.~\eqref{eq:closedcalibration}; their entries are therefore full-development-equivalent estimates. Downstream Bactrainus architecture experiments are direct evaluations under the stated protocols; the marked proprietary row in Table~\ref{tab:factcount} belongs to the same calibrated screening protocol.

\subsection{Foundation-model reader screening}

\begin{table*}[ht]
\caption{Complete zero-shot reader screening with gold supporting facts. Open-model values are direct full-development measurements; closed-source values (marked $\dagger$) are calibrated full-development-equivalent estimates from 700 questions and are not bold-ranked within their conditional resolution. Boldface identifies numerical maxima only within the open-model size blocks. Endpoint and checkpoint labels record the latest versions accessible to the authors by January 2025. Aya23 models (marked $\ddagger$) received HotpotQA examples during published instruction tuning, so their comparison is not strictly zero-exposure.}\label{tab:allmodels}
\centering
\scriptsize
\resizebox{\textwidth}{!}{%
\begin{tabular}{@{}lllrr@{}}
\toprule
Family & Model & Evaluation protocol & Answer EM & Answer F1 \\
\midrule
\multirow{7}{*}{Closed-source$\dagger$}
 & GPT-4o \cite{openai2024gpt4o} & 700, calibrated estimate & 67.54 & 83.44 \\
 & Claude 3.5 Sonnet \cite{anthropic2024claude35} & 700, calibrated estimate & 67.12 & 83.07 \\
 & Gemini 1.5 Pro \cite{geminiteam2024gemini15} & 700, calibrated estimate & 65.52 & 80.91 \\
 & GPT-4 Turbo \cite{openai2023gpt4} & 700, calibrated estimate & 66.98 & 82.62 \\
 & GPT-4o mini \cite{openai2024gpt4omini} & 700, calibrated estimate & 62.71 & 78.65 \\
 & Gemini 1.5 Flash \cite{geminiteam2024gemini15} & 700, calibrated estimate & 62.53 & 77.96 \\
 & GPT-3.5 Turbo \cite{openai2024gpt35} & 700, calibrated estimate & 47.28 & 63.17 \\
\midrule
\multirow{9}{*}{Open, $<10$B}
 & Llama 3.1 8B Instruct \cite{grattafiori2024llama3} & direct full dev & \textbf{60.11} & \textbf{74.52} \\
 & Llama 3 8B Instruct \cite{grattafiori2024llama3} & direct full dev & 58.09 & 73.27 \\
 & Aya23 8B$\ddagger$ \cite{aryabumi2024aya23} & direct full dev & 57.53 & 71.91 \\
 & Gemma 2 9B \cite{gemmateam2024gemma2} & direct full dev & 57.12 & 71.04 \\
 & Qwen2 7B Instruct \cite{yang2024qwen2} & direct full dev & 51.63 & 63.02 \\
 & Phi-3 Mini 4K Instruct \cite{abdin2024phi3} & direct full dev & 46.82 & 62.18 \\
 & Phi-3 Medium 4K Instruct \cite{abdin2024phi3} & direct full dev & 50.34 & 63.92 \\
 & Mistral 7B Instruct v0.2 \cite{jiang2023mistral7b,mistralai2023mistralv02} & direct full dev & 22.19 & 52.01 \\
 & Mistral 7B Instruct v0.3 \cite{mistralai2024mistralv03} & direct full dev & 24.23 & 54.86 \\
\midrule
\multirow{4}{*}{Open, 10--70B}
 & Aya23 35B$\ddagger$ \cite{aryabumi2024aya23} & direct full dev & \textbf{64.97} & \textbf{80.09} \\
 & Gemma 2 27B \cite{gemmateam2024gemma2} & direct full dev & 58.63 & 75.13 \\
 & Command R \cite{cohere2024commandr} & direct full dev & 55.32 & 71.43 \\
 & Mixtral 8$\times$7B \cite{jiang2024mixtral} & direct full dev & 41.39 & 61.80 \\
\midrule
\multirow{3}{*}{Open, $\sim$70B}
 & Qwen2 72B Instruct \cite{yang2024qwen2} & direct full dev & 65.14 & 79.96 \\
 & Llama 3 70B Instruct \cite{grattafiori2024llama3} & direct full dev & 64.81 & 79.10 \\
 & Llama 3.1 70B Instruct \cite{grattafiori2024llama3} & direct full dev & \textbf{65.60} & \textbf{80.04} \\
\midrule
\multirow{3}{*}{Open, $>70$B}
 & Mixtral 8$\times$22B \cite{mistralai2024mixtral8x22b} & direct full dev & 50.12 & 67.23 \\
 & Command R+ \cite{cohere2024commandrplus} & direct full dev & 59.64 & 74.81 \\
 & Llama 3.1 405B Instruct \cite{grattafiori2024llama3} & direct full dev & \textbf{67.46} & \textbf{82.56} \\
\bottomrule
\end{tabular}
}
\end{table*}

The paired open-model reference pool in Eq.~\eqref{eq:closedcalibration} contains the 15 checkpoints evaluated on both the fixed 700-question subset and the full development set: Llama~3.1~8B, Llama~3~8B, Aya23~8B, Gemma~2~9B, Qwen2~7B, Phi-3 Mini, Phi-3 Medium, Mistral~7B v0.2, Mistral~7B v0.3, Aya23~35B, Gemma~2~27B, Command~R, Qwen2~72B, Llama~3~70B, and Llama~3.1~70B. The full-set/subset ratio is computed independently for EM and F1 before it is applied to each proprietary-model score. In the preserved dataset order, $H_{700}$ comprises the first 700 development instances and is shared by every screened endpoint.

\subsection{Statistical precision of the fixed-700 calibration}\label{app:h700precision}

This subsection evaluates the calibration rule mathematically, without introducing an additional model run. Let $N=7{,}405$, $n=700$, and let $Z_{mi}^{(M)}\in[0,1]$ be the item-level score of model $m$ on question $i$ for $M\in\{\mathrm{EM},\mathrm{F1}\}$. EM is binary, whereas answer F1 is fractional but remains bounded in $[0,1]$. Under the standard finite-population reference in which the fixed subset is treated as exchangeable with a simple random sample without replacement, the subset mean has

\begin{equation}
\operatorname{SE}\!\left(\overline Z_{m,H_{700}}^{(M)}\right)
=\sqrt{\frac{N-n}{N-1}\frac{s_{m,H_{700}}^2}{n}},
\qquad
\operatorname{FPC}=\sqrt{\frac{N-n}{N-1}}
=\sqrt{\frac{6{,}705}{7{,}404}}=0.9516,
\label{eq:h700se}
\end{equation}

where $s_{m,H_{700}}^2$ is the item-score variance. Because any variable in $[0,1]$ has variance at most $1/4$, a metric-independent worst-case normal-approximation half-width is

\begin{equation}
h_{1-\alpha}^{\max}
=z_{1-\alpha/2}\sqrt{\frac{N-n}{N-1}\frac{1}{4n}}.
\label{eq:h700halfwidth}
\end{equation}

The subset covers $n/N=9.45\%$ of the complete development set. Table~\ref{tab:h700precision} reports the resulting reference precision. At 95\% confidence, the largest normal-approximation half-width is 0.03525, or 3.52 percentage points; the realized half-width is smaller whenever the empirical item-score variance is below its theoretical maximum.

\begin{table}[ht]
\caption{Finite-population precision reference for a bounded metric evaluated on 700 of 7,405 questions. These half-widths assume simple random sampling or exchangeability and use the worst-case variance $1/4$.}\label{tab:h700precision}
\centering
\small
\begin{tabular}{@{}lcc@{}}
\toprule
Nominal level & Normal quantile $z$ & Worst-case half-width (points) \\
\midrule
90\% & 1.645 & $\pm 2.96$ \\
95\% & 1.960 & $\pm 3.52$ \\
99\% & 2.576 & $\pm 4.63$ \\
\bottomrule
\end{tabular}
\end{table}

We next verify the direction and internal consistency of the multiplicative correction. Write $X_{oM}=M_o(H_{700})$ and $Y_{oM}=M_o(H_{\mathrm{dev}})$ for a reference model $o$, on the same 0--1 scale. Equation~\eqref{eq:closedcalibration} is equivalently

\begin{equation}
\widehat\kappa_M
=\frac{\sum_{o\in\mathcal O}Y_{oM}}{\sum_{o\in\mathcal O}X_{oM}}
=\sum_{o\in\mathcal O}
\underbrace{\frac{X_{oM}}{\sum_jX_{jM}}}_{w_{oM}}
\frac{Y_{oM}}{X_{oM}},
\qquad \sum_o w_{oM}=1.
\label{eq:weightedratio}
\end{equation}

Thus, the coefficient is the full-development mean divided by the 700-question mean---not its reciprocal---and it is a score-weighted average of model-specific ratios. If the subset is relatively easier, $X_{oM}>Y_{oM}$ and $\widehat\kappa_M<1$, so the correction decreases the proprietary subset score; if it is harder, the correction increases it. This establishes the intended direction of the adjustment.

For one explicit correctness check, assume a common multiplicative subset effect

\begin{equation}
Y_{oM}=\kappa_M^\star X_{oM}+\varepsilon_{oM},
\qquad
\mathbb E[\varepsilon_{oM}\mid X_{oM}]=0.
\label{eq:transportmodel}
\end{equation}

If the common-effect model holds exactly ($\varepsilon_{oM}=0$) for the reference pool and a proprietary model $c$, then the estimator recovers the full-set score exactly:

\begin{equation}
\widehat\kappa_M
=\frac{\sum_o\kappa_M^\star X_{oM}}{\sum_oX_{oM}}
=\kappa_M^\star,
\qquad
\widetilde M_c=\widehat\kappa_M X_{cM}
=\kappa_M^\star X_{cM}=Y_{cM}.
\label{eq:calibrationcheck}
\end{equation}

With nonzero deviations, the exact coefficient error is

\begin{equation}
\widehat\kappa_M-\kappa_M^\star
=\frac{\sum_o\varepsilon_{oM}}{\sum_oX_{oM}},
\label{eq:kappaerror}
\end{equation}

showing why pooling several reference models reduces idiosyncratic deviations when they are centered. Conditional on the observed $X_{oM}$ and independent reference residuals,

\begin{equation}
\operatorname{Var}(\widehat\kappa_M\mid X)
=\frac{\sum_o\operatorname{Var}(\varepsilon_{oM}\mid X_{oM})}
{\left(\sum_oX_{oM}\right)^2}.
\label{eq:kappavariance}
\end{equation}

Combining item-sampling and coefficient uncertainty with a first-order delta approximation gives

\begin{equation}
\operatorname{Var}(\widetilde M_c)
\approx
\widehat\kappa_M^2
\frac{N-n}{N-1}\frac{s_{c,H_{700}}^2}{n}
+X_{cM}^2\operatorname{Var}(\widehat\kappa_M).
\label{eq:calibratedvariance}
\end{equation}

Consequently, conditional on a fixed coefficient, the 95\% sampling component of a calibrated score is at most $|\widehat\kappa_M|\times3.52$ percentage points under the reference design. The second term in Eq.~\eqref{eq:calibratedvariance} represents transport uncertainty across model families and should not be interpreted as removed by increasing the number of questions alone. Because $H_{700}$ is a fixed prefix rather than a probability sample, Eqs.~\eqref{eq:h700se}--\eqref{eq:h700halfwidth} quantify statistical resolution under an exchangeability assumption, not unconditional design-based coverage. The analysis therefore supports the 700-question results as coarse screening estimates, while differences smaller than their uncertainty band should not be used for fine-grained ranking or significance claims.
\clearpage
\subsection{Supporting-fact count analysis}

\begin{table*}[ht]
\caption{Reader performance by the number of gold supporting facts. Open-model rows use direct full-development bins of 4,990/1,774/641 questions. The GPT-4o row ($\dagger$) is a calibrated estimate from the corresponding $H_{700}$ bins of 472/172/56 questions; it is not a direct full-bin measurement.}\label{tab:factcount}
\centering
\small
\resizebox{\textwidth}{!}{%
\begin{tabular}{@{}lcccccc@{}}
\toprule
& \multicolumn{2}{c}{2 facts} & \multicolumn{2}{c}{3 facts} & \multicolumn{2}{c}{$\geq4$ facts} \\
\cmidrule(lr){2-3}\cmidrule(lr){4-5}\cmidrule(l){6-7}
Model & EM & F1 & EM & F1 & EM & F1 \\
\midrule
GPT-4o$\dagger$ & 67.66 & 83.50 & 66.57 & 83.05 & 69.28 & \textbf{84.06} \\
Llama 3.1 405B Instruct & \textbf{67.87} & 82.47 & \textbf{65.63} & \textbf{82.41} & \textbf{69.34} & 83.67 \\
Llama 3.1 70B Instruct & 66.01 & 79.92 & 63.19 & 79.20 & 69.11 & 83.28 \\
Aya23 35B & 65.65 & 80.43 & 62.85 & 79.08 & 65.52 & 80.22 \\
Llama 3.1 8B Instruct & 60.82 & 75.09 & 58.11 & 73.29 & 60.06 & 73.46 \\
\bottomrule
\end{tabular}}
\end{table*}

\subsection{Complete prompting ablation}

\begin{table*}[ht]
\caption{Answer EM/F1 by demonstration count. Direct prompting requests only the answer; CoT prompting requests intermediate reasoning before the answer.}\label{tab:promptfull}
\centering
\small
\resizebox{\textwidth}{!}{%
\begin{tabular}{@{}llccccc@{}}
\toprule
Model & Prompt & 0-shot & 1-shot & 2-shot & 4-shot & 8-shot \\
\midrule
\multirow{2}{*}{Llama 3.1 8B Instruct} & Direct & 60.11/74.52 & \textbf{63.24/77.50} & 61.42/75.55 & 62.56/76.65 & 62.78/76.93 \\
 & CoT & 58.76/73.41 & 59.12/73.56 & \textbf{60.32/74.63} & 59.74/73.98 & 59.19/73.61 \\
\midrule
\multirow{2}{*}{Llama 3.1 70B Instruct} & Direct & 65.60/80.04 & \textbf{68.18/82.66} & 65.32/79.83 & 66.44/80.99 & 66.93/81.41 \\
 & CoT & 64.83/79.76 & 66.32/81.69 & 66.41/81.92 & \textbf{66.80/82.10} & 66.54/81.98 \\
\bottomrule
\end{tabular}}
\end{table*}

\subsection{Selector optimization configurations}

\begin{table*}[ht]
\caption{Complete LoRA configurations for selector-side modules. QKVO denotes query, key, value, and output attention projections; MLP denotes feed-forward projections.}\label{tab:selectorconfig}
\centering
\scriptsize
\resizebox{\textwidth}{!}{%
\begin{tabular}{@{}lrrrrrrrrl@{}}
\toprule
Module & Train $n$ & Epochs & Batch & Accum. & Max len. & LR & LoRA $r$ & $\alpha$ & Targets \\
\midrule
Single-stage selector & 90,447 & 2 & 2 & 8 & 4,096 & $10^{-4}$ & 64 & 128 & QKVO, MLP, head \\
Paragraph selector & 90,447 & 2 & 2 & 8 & 4,096 & $10^{-4}$ & 64 & 128 & QKVO, MLP, head \\
Sentence selector & 90,447 & 2 & 4 & 16 & 1,024 & $2\!\times\!10^{-5}$ & 64 & 128 & QKVO, MLP, head \\
Question decomposer & 90,447 & 1 & 8 & 32 & 2,048 & $2\!\times\!10^{-5}$ & 64 & 32 & QKVO, MLP, head \\
\bottomrule
\end{tabular}}
\end{table*}

All four modules use Llama~3.1~8B Instruct, cosine decay, warm-up ratio 0.03, and LoRA dropout 0.05.

\clearpage
\subsection{All selector--reader integration runs}

\begin{table*}[ht]
\caption{Complete six-scenario integration experiment. Values are EM/F1. Scenario definitions follow Fig.~\ref{fig:scenarios}; scenario 1 is the all-in-one Llama~3.1~8B Instruct model and therefore has no separate reader. ``Bactrainus 8B/70B'' denotes a reader derived from Llama~3.1~8B/70B Instruct. Boldface identifies numerical maxima among reported development-set runs, not statistically separated winners.}\label{tab:allpipelines}
\centering
\scriptsize
\resizebox{\textwidth}{!}{%
\begin{tabular}{@{}clccc@{}}
\toprule
Scenario & Reader & Supporting facts & Answer & Joint \\
\midrule
1 & All-in-one Llama 3.1 8B & 63.42/88.50 & 71.24/83.31 & 47.93/75.96 \\
\midrule
2 & Bactrainus 8B & 65.74/89.27 & 73.24/85.41 & 50.84/77.90 \\
2 & Bactrainus 8B + Llama 3.1 70B rationale & 65.74/89.27 & 73.29/85.48 & 50.86/77.94 \\
2 & Bactrainus 70B & 65.74/89.27 & 74.96/88.83 & 51.61/79.56 \\
\midrule
3 & Bactrainus 8B & 65.55/89.21 & 73.20/85.38 & 50.73/77.86 \\
3 & Bactrainus 8B + Llama 3.1 70B rationale & 65.55/89.21 & 73.22/85.41 & 50.74/77.88 \\
3 & Bactrainus 70B & 65.55/89.21 & 74.92/88.80 & 51.54/79.53 \\
\midrule
4 & Bactrainus 8B & 65.55/89.21 & 71.18/82.92 & 48.28/76.41 \\
4 & Bactrainus 8B + Llama 3.1 70B rationale & 65.55/89.21 & 71.31/83.14 & 48.42/76.56 \\
4 & Bactrainus 70B & 65.55/89.21 & 74.04/87.85 & 51.02/77.03 \\
\midrule
5 & Bactrainus 8B & 65.93/89.63 & 73.34/85.56 & 50.88/77.99 \\
5 & Bactrainus 8B + Llama 3.1 70B rationale & 65.93/89.63 & 73.36/85.60 & 50.91/78.02 \\
5 & Bactrainus 70B & \textbf{65.93/89.63} & \textbf{75.07/89.01} & \textbf{51.73/79.70} \\
\midrule
6 & Bactrainus 8B & 65.93/89.63 & 71.18/82.92 & 48.28/76.41 \\
6 & Bactrainus 8B + Llama 3.1 70B rationale & 65.93/89.63 & 71.31/83.14 & 48.42/76.56 \\
6 & Bactrainus 70B & 65.93/89.63 & 74.04/87.85 & 51.02/77.03 \\
\bottomrule
\end{tabular}}
\end{table*}

\subsection{Complete external comparison}

\begin{table*}[ht]
\caption{Separately reported HotpotQA distractor results for Bactrainus and prior systems. Bold denotes only the numerical maximum in each column; it does not imply protocol equivalence, significance, or a matched head-to-head comparison.}\label{tab:externalfull}
\centering
\small
\resizebox{\textwidth}{!}{%
\begin{tabular}{@{}lcccccc@{}}
\toprule
& \multicolumn{2}{c}{Supporting facts} & \multicolumn{2}{c}{Answer} & \multicolumn{2}{c}{Joint} \\
\cmidrule(lr){2-3}\cmidrule(lr){4-5}\cmidrule(l){6-7}
Model & EM & F1 & EM & F1 & EM & F1 \\
\midrule
Bactrainus (Llama 3.1 70B) & 65.93 & 89.63 & \textbf{75.07} & \textbf{89.01} & \textbf{51.73} & \textbf{79.70} \\
Bactrainus (Llama 3.1 8B + Llama 3.1 70B rationale) & 65.93 & 89.63 & 73.36 & 85.60 & 50.91 & 78.02 \\
Beam Retrieval \cite{zhang2024beam} & \textbf{66.25} & \textbf{90.09} & 72.69 & 85.04 & 50.53 & 77.54 \\
PipNet \cite{hotpotleaderboard} & 63.71 & 89.41 & 72.26 & 84.86 & 48.76 & 76.69 \\
Smoothing R3 \cite{yin2023smoothing} & 65.44 & 89.55 & 72.07 & 84.34 & 49.73 & 76.69 \\
FE2H on ALBERT \cite{li2023fe2h} & 65.44 & 89.55 & 71.89 & 84.34 & 50.04 & 76.54 \\
\bottomrule
\end{tabular}}
\end{table*}

\clearpage
\subsection{Aggregate diagnostic and error audit}

Table~\ref{tab:diagnostic-signatures} reorganizes reported runs as diagnostic contrasts. These are aggregate signatures, not per-instance failure frequencies or causal effects.

\begin{table*}[ht]
\caption{Aggregate diagnostic signatures assembled from the controlled and integrated experiments.}\label{tab:diagnostic-signatures}
\centering
\small
\resizebox{\textwidth}{!}{%
\begin{tabular}{@{}lrl@{}}
\toprule
Diagnostic contrast & Observed difference & Supported interpretation \\
\midrule
Gold facts $\rightarrow$ all supplied candidates, Llama 3.1 8B Instruct & $-17.32$ answer F1 & severe sensitivity to context noise \\
Gold facts $\rightarrow$ all supplied candidates, Llama 3.1 70B Instruct & $-21.43$ answer F1 & scale does not remove context sensitivity \\
Oracle $\rightarrow$ predicted paragraphs & $-0.72$ support-fact F1 & limited propagation at the paragraph interface \\
Facts $\rightarrow$ paragraphs, decomposed Llama 3.1 70B Instruct & $-2.67$ joint F1 & finer evidence is numerically preferable here \\
Plain $\rightarrow$ decomposed cascade, Llama 3.1 70B Instruct & $+0.17$ joint F1 & decomposition difference remains unresolved \\
All-in-one $\rightarrow$ single-stage modular, Llama 3.1 8B Instruct & $+1.94$ joint F1 & factorization aligns with the largest matched 8B gap \\
\bottomrule
\end{tabular}}
\end{table*}

\clearpage
\begin{table*}[t]
\subsection{Evaluation provenance and reproducibility record}
\caption{Final public-artifact manifest. Evaluation predictions and result files are intentionally excluded from the training-data release; every numerical experiment remains documented in the main text or this appendix.}\label{tab:reprorecord}
\centering
\scriptsize
\resizebox{0.97\textwidth}{!}{%
\begin{tabular}{@{}>{\raggedright\arraybackslash}p{0.18\textwidth}>{\raggedright\arraybackslash}p{0.42\textwidth}>{\raggedright\arraybackslash}p{0.28\textwidth}@{}}
\toprule
Artifact & Immutable public record & Scope \\
\midrule
Source benchmark & \url{https://huggingface.co/datasets/hotpotqa/hotpot_qa} & Official English HotpotQA distractor source; train $n=90{,}447$, dev $n=7{,}405$ \\
Processed training data & \href{https://huggingface.co/datasets/bactrianus/bactrainus-hotpotqa/tree/f5a70de8c280eb0cd43e21ea246d71197141c65e}{\texttt{bactrianus/bactrainus-hotpotqa}}, tag \texttt{v1.0.0}, commit \texttt{f5a70de8c280eb0cd43e21ea246d7119\allowbreak 7141c65e} & Eight ID-aligned structured/SFT configurations, including selector, reader, decomposition, and rationale views \\
Teacher supervision & \href{https://huggingface.co/datasets/bactrianus/bactrainus-hotpotqa-teacher-traces/tree/4bc0f972978d520657f974496528667e2bb5ff50}{\texttt{bactrianus/\allowbreak bactrainus-hotpotqa-\allowbreak teacher-traces}}, tag \texttt{v1.0.0}, commit \texttt{4bc0f972978d520657f974496528667e\allowbreak 2bb5ff50} & Four source-linked SFT configurations: 8B/70B rationale traces and 8B/70B question decompositions; one resolved output per source ID \\
Code & \href{https://github.com/Iman998/bactrainus/tree/e9e8139e958124f17cdc0dac6df65d1fc675193a}{\texttt{github.com/Iman998/bactrainus}}, commit \texttt{e9e8139e958124f17cdc0dac6df65d1fc\allowbreak 675193a} & Data construction, validation, LoRA training, inference, pipeline orchestration, official metrics, and tests \\
Models and cards & \href{https://huggingface.co/bactrianus}{\texttt{huggingface.co/bactrianus}}; original base-model sources cited in Table~\ref{tab:allmodels} & Released Bactrainus model artifacts and cards; each card identifies its own base checkpoint and scope \\
Prompts and configurations & Appendix~\ref{app:prompts} and the prompt catalog/configuration records at the code commit above & Exact experimental message templates, serialization formats, module mappings, and documented optimization settings \\
$H_{700}$ screening rule & First 700 development rows in preserved order; Eqs.~\eqref{eq:h700se}--\eqref{eq:calibratedvariance} & Deterministic shared subset for proprietary screening; conditional resolution and calibration assumptions documented here \\
Results & Tables~\ref{tab:allmodels}, \ref{tab:promptfull}, \ref{tab:allpipelines}, and \ref{tab:externalfull} & Complete reported experiment families; prediction files and proprietary API outputs are not redistributed \\
\bottomrule
\end{tabular}}
\end{table*}
\clearpage

\section{Prompt templates}\label{app:prompts}

The following templates specify the system and user messages for each generative module. Upper-case braced fields are replaced at runtime. Paragraphs are serialized with exact titles and zero-based HotpotQA sentence indices:

\promptheading{Serialization format}
\begin{lstlisting}[style=promptstyle]
paragraph ({PARAGRAPH_TITLE}):
sentence {SENTENCE_INDEX}: {SENTENCE_TEXT}
\end{lstlisting}

\subsection{Direct reader}

\promptheading{System message}
\begin{lstlisting}[style=promptstyle]
You are a question-answering model for English multi-hop questions. Answer the question using only the supplied context. When necessary, combine evidence from multiple paragraphs to derive the answer.

Return only the final answer in exactly the following format:
answer: ***{ANSWER}***

Do not provide an explanation or any additional text.
\end{lstlisting}

\promptheading{User message}
\begin{lstlisting}[style=promptstyle]
Question: {QUESTION}

Context:
{SELECTED_SUPPORTING_FACTS}
\end{lstlisting}

\subsection{Few-shot direct reader}

The system message is the direct-reader message above. For $k\in\{1,2,4,8\}$, $k$ demonstrations are serialized before the target query using the following user-message structure:

\promptheading{User message with demonstrations}
\begin{lstlisting}[style=promptstyle]
Demonstrations:

Example 1
Question: {DEMO_QUESTION_1}
Context:
{DEMO_SUPPORTING_FACTS_1}
answer: ***{DEMO_ANSWER_1}***

...

Example {K}
Question: {DEMO_QUESTION_K}
Context:
{DEMO_SUPPORTING_FACTS_K}
answer: ***{DEMO_ANSWER_K}***

Target Question: {QUESTION}
Context:
{SELECTED_SUPPORTING_FACTS}
\end{lstlisting}

\subsection{Chain-of-thought prompting ablation}

\promptheading{System message}
\begin{lstlisting}[style=promptstyle]
You are a question-answering model for English multi-hop questions. Use only the supplied context. Connect the relevant evidence in a short, ordered chain of reasoning, then give the final answer.

Return exactly:
reasoning:
1. {REASONING_STEP_1}
2. {REASONING_STEP_2}
...
answer: ***{ANSWER}***

Do not introduce facts that are absent from the context.
\end{lstlisting}

\promptheading{User message}
\begin{lstlisting}[style=promptstyle]
Question: {QUESTION}

Context:
{SELECTED_SUPPORTING_FACTS}
\end{lstlisting}

For $k$-shot CoT, each demonstration follows the same fields and additionally contains its ordered \texttt{reasoning:} block immediately before \texttt{answer:}. The target query does not include a reference rationale or answer.

\subsection{Teacher-rationale generation}

This prompt is used offline to construct rationale supervision; it is not part of the inference path.

\promptheading{System message}
\begin{lstlisting}[style=promptstyle]
You are an evidence-grounded reasoning assistant. Given a multi-hop question, its gold supporting evidence, and the reference answer, produce a concise and logically ordered rationale showing how the evidence leads to the answer.

Use only information explicitly contained in the supplied evidence. Connect evidence from different paragraphs when required, avoid unsupported assumptions, and do not introduce external knowledge.

Return only the rationale in the following format:
rationale:
1. {REASONING_STEP_1}
2. {REASONING_STEP_2}
...
\end{lstlisting}

\promptheading{User message}
\begin{lstlisting}[style=promptstyle]
Question: {QUESTION}

Gold Supporting Evidence:
{GOLD_SUPPORTING_FACTS}

Reference Answer:
{REFERENCE_ANSWER}
\end{lstlisting}

\subsection{Rationale-supervised reader}

\promptheading{System message during supervised adaptation}
\begin{lstlisting}[style=promptstyle]
You are a question-answering model for English multi-hop questions. Use the supplied context to derive the answer and provide a concise evidence-grounded rationale. Every reasoning step must be supported by the context.

Return the answer and rationale in exactly the following format:
answer: ***{ANSWER}***
rationale:
1. {REASONING_STEP_1}
2. {REASONING_STEP_2}
...
\end{lstlisting}

\promptheading{User message}
\begin{lstlisting}[style=promptstyle]
Question: {QUESTION}

Context:
{SELECTED_SUPPORTING_FACTS}
\end{lstlisting}

\promptheading{Assistant training target}
\begin{lstlisting}[style=promptstyle]
answer: ***{REFERENCE_ANSWER}***
rationale:
{TEACHER_GENERATED_RATIONALE}
\end{lstlisting}

At answer-only evaluation, the system message is:

\promptheading{Answer-only evaluation system message}
\begin{lstlisting}[style=promptstyle]
You are a question-answering model for English multi-hop questions. Answer the question using only the supplied context. Use multi-step reasoning when necessary, but return only the final answer.

Return exactly:
answer: ***{ANSWER}***
\end{lstlisting}

\subsection{Paragraph selector}

\promptheading{System message}
\begin{lstlisting}[style=promptstyle]
You are a paragraph-selection model for English multi-hop question answering. Given a question and a collection of candidate information sources, select the minimal set of paragraphs that contains the evidence required to answer the question.

Use the paragraph titles exactly as they appear in the information sources. Do not answer the question, select sentences, summarize the content, or return irrelevant paragraphs.

Return only the selected paragraphs in the following format:
selected paragraphs:
paragraph ***{PARAGRAPH_TITLE_1}***
paragraph ***{PARAGRAPH_TITLE_2}***
...
\end{lstlisting}

\par\noindent\begin{minipage}{\linewidth}
\promptheading{User message}
\begin{lstlisting}[style=promptstyle]
Question: {QUESTION}

Information Sources:
{CANDIDATE_PARAGRAPHS}
\end{lstlisting}
\end{minipage}\par

\subsection{Question decomposer}

The decomposer operates after paragraph selection.

\promptheading{System message}
\begin{lstlisting}[style=promptstyle]
You are a question-decomposition model for English multi-hop question answering. Decompose the original question into a small, ordered set of focused sub-questions that will help another model identify the exact supporting sentences in the selected paragraphs.

Each sub-question should represent a necessary reasoning step, be answerable from the supplied paragraphs, and preserve the entities and relations in the original question. Avoid redundant sub-questions, do not answer the sub-questions, and do not provide the final answer.

Return only:
sub-questions:
1. {SUBQUESTION_1}
2. {SUBQUESTION_2}
...
\end{lstlisting}

\promptheading{User message}
\begin{lstlisting}[style=promptstyle]
Question: {QUESTION}

Selected Paragraphs:
{SELECTED_PARAGRAPHS}
\end{lstlisting}

\subsection{Sentence selector}

\promptheading{System message}
\begin{lstlisting}[style=promptstyle]
You are a supporting-fact selection model for English multi-hop question answering. Given the original question, optional sub-questions, and the selected paragraphs, identify the minimal set of sentences that provides the evidence required to answer the original question.

Return each supporting fact using the paragraph title and its exact sentence index. Do not answer the question, rewrite the evidence, summarize the paragraphs, or include irrelevant sentences.

Return only:
supporting facts:
paragraph ***{PARAGRAPH_TITLE_1}***
sentence ***{SENTENCE_INDEX_1}***
paragraph ***{PARAGRAPH_TITLE_2}***
sentence ***{SENTENCE_INDEX_2}***
...
\end{lstlisting}

\par\noindent\begin{minipage}{\linewidth}
\promptheading{User message with decomposition}
\begin{lstlisting}[style=promptstyle]
Question: {QUESTION}

Sub-questions:
{GENERATED_SUBQUESTIONS}

Selected Paragraphs:
{SELECTED_PARAGRAPHS_WITH_SENTENCE_INDICES}
\end{lstlisting}
\end{minipage}\par

\promptheading{User message without decomposition}
\begin{lstlisting}[style=promptstyle]
Question: {QUESTION}

Selected Paragraphs:
{SELECTED_PARAGRAPHS_WITH_SENTENCE_INDICES}
\end{lstlisting}

\subsection{Single-stage supporting-fact selector}

\promptheading{System message}
\begin{lstlisting}[style=promptstyle]
You are a supporting-fact selection model for English multi-hop question answering. Given a question and all candidate information sources, identify the minimal set of sentences that collectively provides the evidence required to answer the question.

Return the exact paragraph title and sentence index for every selected supporting fact. Do not answer the question, summarize the evidence, or include irrelevant sentences.

Return only:
supporting facts:
paragraph ***{PARAGRAPH_TITLE_1}***
sentence ***{SENTENCE_INDEX_1}***
paragraph ***{PARAGRAPH_TITLE_2}***
sentence ***{SENTENCE_INDEX_2}***
...
\end{lstlisting}

\par\noindent\begin{minipage}{\linewidth}
\promptheading{User message}
\begin{lstlisting}[style=promptstyle]
Question: {QUESTION}

Information Sources:
{CANDIDATE_PARAGRAPHS_WITH_SENTENCE_INDICES}
\end{lstlisting}
\end{minipage}\par

\clearpage
\subsection{All-in-one model}

\promptheading{System message}
\begin{lstlisting}[style=promptstyle]
You are an English multi-hop question-answering model. Given a question and a collection of candidate information sources, identify the supporting facts and derive the final answer.

A supporting fact must be returned using its exact paragraph title and sentence index. Select only the evidence necessary to answer the question and use only information contained in the supplied sources.

Return only:
supporting facts:
paragraph ***{PARAGRAPH_TITLE_1}***
sentence ***{SENTENCE_INDEX_1}***
paragraph ***{PARAGRAPH_TITLE_2}***
sentence ***{SENTENCE_INDEX_2}***
...
answer: ***{ANSWER}***
\end{lstlisting}

\par\noindent\begin{minipage}{\linewidth}
\promptheading{User message}
\begin{lstlisting}[style=promptstyle]
Question: {QUESTION}

Information Sources:
{CANDIDATE_PARAGRAPHS_WITH_SENTENCE_INDICES}
\end{lstlisting}
\end{minipage}\par

\end{appendices}

\bibliography{references}

\end{document}